\documentclass[letterpaper, 10 pt, conference]{ieeeconf}  

\IEEEoverridecommandlockouts                              

\usepackage{hyperref}
\hypersetup{bookmarksopen,bookmarksnumbered,
pdfpagemode=UseOutlines,
colorlinks=true,
linkcolor=blue,
anchorcolor=blue,
citecolor=blue,
filecolor=blue,
menucolor=blue,
urlcolor=blue
}
\usepackage{upgreek}
\usepackage{amsfonts,amssymb}
\usepackage{bm}
\usepackage{bbm}

\usepackage{amsthm} 
\usepackage{thmtools}
\usepackage{mathtools}
\usepackage{xspace}
\usepackage{booktabs} 
\usepackage{tabulary}
\usepackage[usenames,dvipsnames,table]{xcolor} 
\usepackage{diagbox}
\usepackage[ruled,vlined,linesnumbered]{algorithm2e}  
\SetKwComment{Comment}{$\triangleright$\ }{}
\usepackage{ifthen,version}
\usepackage{soul}
\usepackage{csquotes}
\usepackage{microtype}      
\usepackage{lipsum}
\usepackage{tikz}
\let\labelindent\relax
\usepackage{todonotes}
\usepackage{enumitem}

\usepackage{amsthm} 
\usepackage{thmtools}
\usepackage{multirow}
\usepackage{amsfonts}
\usepackage{booktabs}
\usepackage{siunitx}
\usepackage{mathtools}
\usepackage{epstopdf}
\usepackage{xspace}
\usepackage{cleveref}
\usepackage{booktabs}       
\usepackage{subcaption}
\usepackage{comment}
\usepackage{graphicx}
\newcommand{\method}[1]{PACT}

\newcommand{\smnote}[1]{{\xxnote{SM}{blue}{#1}}} 
\newcommand{\xxnote}[3]{}
\ifx\hidenotes\undefined
  \usepackage{color}
  \renewcommand{\xxnote}[3]{\color{#2}{#1: #3}}
\fi

\graphicspath{{figs/}}
\usepackage{wrapfig}

\usepackage[noadjust]{cite}

\usepackage[normalem]{ulem}
\newcommand\redsout{\bgroup\markoverwith{\textcolor{red}{\rule[0.5ex]{2pt}{0.7pt}}}\ULon}

\usepackage{multirow}
\usepackage{pbox}

\newtheoremstyle{hypstyle}
{3pt} 
{3pt} 
{\itshape} 
{} 
{\bfseries} 
{.} 
{.5em} 
{} 

\theoremstyle{hypstyle}

\title{\LARGE \bf
PACT: Perception-Action Causal Transformer \\for Autoregressive Robotics Pre-Training
\vspace{3mm} 

\author{Rogerio Bonatti, Sai Vemprala, Shuang Ma, Felipe Frujeri, Shuhang Chen, Ashish Kapoor\\
Microsoft}
\vspace{-6mm}
}

\begin{document}

\maketitle
\thispagestyle{plain}
\pagestyle{plain}



\begin{abstract}

Robotics has long been a field riddled with complex systems architectures whose modules and connections, whether traditional or learning-based, require significant human expertise and prior knowledge.
Inspired by large pre-trained language models, this work introduces a paradigm for pre-training a general purpose representation that can serve as a starting point for multiple tasks on a given robot. 
We present the Perception-Action Causal Transformer (\method{}), a generative transformer-based architecture that aims to build representations directly from robot data in a self-supervised fashion. 
Through autoregressive prediction of states and actions over time, our model implicitly encodes dynamics and behaviors for a particular robot.
Our experimental evaluation focuses on the domain of mobile agents, where we show that this robot-specific representation can function as a single starting point to achieve distinct tasks such as safe navigation, localization and mapping. 
We evaluate two form factors: a wheeled robot that uses a LiDAR sensor as perception input (MuSHR), and a simulated agent that uses first-person RGB images (Habitat). 
We show that finetuning small task-specific networks on top of the larger pretrained model results in significantly better performance compared to training a single model from scratch for all tasks simultaneously, and comparable performance to training a separate large model for each task independently.
By sharing a common good-quality representation across tasks we can lower overall model capacity and speed up the real-time deployment of such systems.
    
\end{abstract}

\section{Introduction}

Recent advances in machine learning architectures have started a paradigm shift from task-specific models towards large general purpose models. Such a shift has most commonly been observed in the domain of natural language, as evidenced by large language models such as BERT \cite{devlin2018bert}, GPT-3 \cite{brown2020language} and Megatron-Turing \cite{smith2022using}, as well as in computer vision~\cite{yuan2021florence, su2019vlbert, lu2019vilbert}. 

Some of these large models already combine multiple data modalities such as text, images, video, audio, as well as the relationship between datapoints over time~\cite{jaegle2021perceiver,akbari2021vatt}. The use of foundational models is appealing because they are trained on broad datasets over a wide variety of downstream tasks, and therefore provide general skills which can be used directly or with minimal fine-tuning to new applications. More recently, large pretrained models have been applied to multi-task learning spanning multiple domains \cite{gato}.

While machine learning models are finding widespread use in robotics, most of them have been task or hardware-specific, which necessitate redesign and retraining if there are minor changes in robot dynamics, environment, or operational objectives~\cite{kaufmann2018deep}. 
A contrast can be drawn between such approaches in robotics and research in the domain of natural language. For instance, hand-crafted language models that encoded grammatical rules and syntax have been replaced by large models that can learn directly from data. 
Such large models often encode general purpose information about language, grammar, and can be finetuned for specific tasks with relative ease. 
Similarly, we envisage a general architecture for robotics that requires less priors and domain expertise, and can serve as a starting point for several tasks. 

The representation learning methods presented in this paper are built to be agnostic to the specific robotics domain as long as we use states and actions to represent our system. 
However, this paper focuses on domains related to mobile agents, where the typical robot autonomy pipeline involves objectives such as localization, mapping, and planning.

\begin{figure}
    \centering
    \includegraphics[width=1.00\columnwidth]{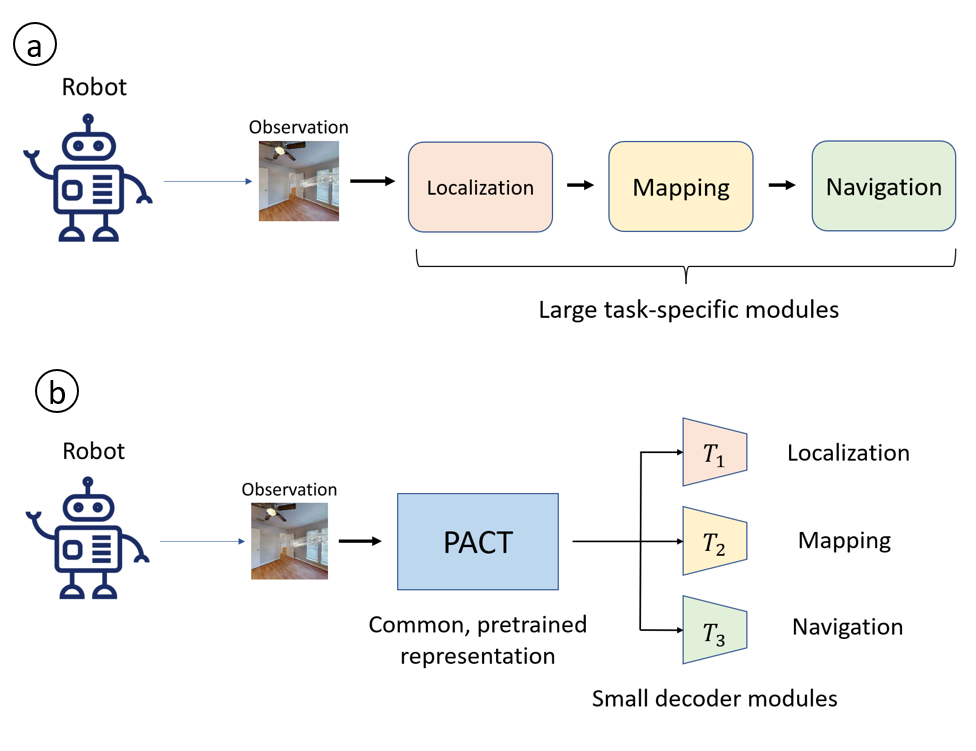}
    \caption{\small{Comparison between a) traditional robotics architecture with hand-crafted modules, and b) foundation model architecture with common pre-trained representations and lightweight downstream modules.}}
    \label{fig:main_fig}
    \vspace{-6mm}
\end{figure}

Recent works have just started to explore the use of pretrained models towards robotic decision-making.
The main challenges in robotics that differ from traditional language, vision and vision-language models are:

\textbf{Multi-modal data: } unlike large language models that are limited to a single domain of data, robotics models often need to process disparate input modalities (images, point clouds, velocities, arbitrary features),  and output low and high-level decisions;

\textbf{Sequential decision-making: } actions have consequences, and a robotics model must be able to not only summarize the current information, but also reason about the multitude of possible futures scenarios and the causal relationships between the problem’s variables;

\textbf{Expensive and scarce data: }  unlike language and vision data, real-world robotics data is prohibitively expensive to collect in large quantities, and therefore requires the use of simulators which often involves a sim-to-real gap. In addition, data collected for a particular robot form does not necessarily generalize across new platforms or tasks, and sufficient data diversity is problematic.

In this paper, we present a pre-training paradigm for robotics, {with special experimental focus on the domain of mobile agents}. 
We identify that at their core, most robotic agents process a perception-action loop between their states/observations, and associated actions - and that an understanding of state-action transitions is beneficial for several tasks in the pipeline of robot autonomy. We hence present the Perception-Action Causal Transformer (PACT), a transformer-based generative model that is trained on sequences of states and actions coming from robot trajectories. 
By learning to autoregressively predict such sequences, \method{} implicitly encodes general purpose information such as the notion of which next state would be reached from a current state and action (robot dynamics), as well as the notion of which action to take given a certain state. 
{We train \method{} on trajectories obtained from two different navigation domains (MuSHR wheeled car and virtual Habitat agent)}, and we show that representations learnt by \method{} can act as a base to efficiently solve several robotics tasks such as localization, mapping and navigation. 

Our main contributions are listed below:

\begin{itemize}
    \item {We propose a pretraining architecture for robotics tasks based on a Perception-Action Causal Transformer (\method{}), which uses an autoregressive objective to encode state-action transitions;}
    \item {Within the domain of mobile agents, we show that this pretrained model can provide a single reasonable starting point for the tasks of  safe navigation, localization, and mapping. We experimentally verify that our frozen common representation can achieve performance levels similar to training separate networks of equal capacity individually for each downstream task, and significantly higher performance than training a single network for all tasks from scratch.}
    \item {We provide experiments in navigation tasks that span different data modalities (LiDAR and RGB) and robot dynamics models (wheeled, omni-directional), and real-world robot validation.}
\end{itemize}

\section{Related Work} 
\label{sec:related_work}

\textbf{Learning robotics representations:}
Approaches for task learning in robotics can be categorized into three main segments. The first, and the most common, is the task-specific training approach where modules are designed specific to each task. Such methods have been proposed for several of the most common robotic tasks such as visual/LiDAR based localization, mapping, path planning and control ~\cite{li2019net, barsan2020learning}. Another category involves multi-task learning approaches, where models are trained jointly to be able to solve several tasks~\cite{kalashnikov2021mt, huang2021generalization}. Finally, there exists a class of techniques that perform task-agnostic pre-training, whose representations can later be finetuned for a task of choice~\cite{parisi2022unsurprising,ma2022compass,nair2022r3m}. 

\textbf{Multi-modal robotics representations:}
Representation learning is a rapidly growing field. 
The existing visual-language representation approaches primarily rely on BERT-style~\cite{devlin2018bert} training objectives to model the cross-modal alignments. 
Common downstream tasks consist of visual question-answering, grounding, retrieval and captioning etc.~\cite{sun2019videobert, lu2019vilbert, zhou2020unified, su2019vlbert}. 
Learning representations for robotics tasks poses additional challenges, as perception data is conditioned on the motion policy and model dynamics~\cite{bommasani2021opportunities}.
Visual-language navigation of embodied agents is well-established field with clear benchmarks and simulators~\cite{szot2021habitat,Anderson2018room2room}, and multiple works explore the alignment of vision and language data by combining pre-trained models with fine-tuning~\cite{hao2020genericVLN,thomason2020vision,nguyen2019helpAnna}
To better model the visual-language alignment, \cite{ma2019progressestimation} also proposed a co-grounding attention mechanism.
In the manipulation domain we also find the work of \cite{shridhar2022cliport}, which uses CLIP~\cite{radford2021learning} embeddings to combine semantic and spatial information.
%

\textbf{Transformers in robotics:}
Transformers were originally introduced in the language processing domain~\cite{vaswani2017attention}, but quickly proved to be useful in modeling long-range data dependencies other domains.
Within robotics we see the first transformers architectures being used for trajectory forecasting~\cite{giuliari2021transformer}, motion planning~\cite{bucker2022reshaping,chaplot2021differentiable}, and reinforcement learning~\cite{chen2021decision,janner2021offline}.
Our main difference between the related works in \cite{chen2021decision,janner2021offline} (Decision Transformer and Trajectory Transformer) is that they are focused on training a model for a single task, while we propose learning representations amenable to multiple downstream tasks for a robot.


\section{Pretraining Approach}
\label{sec:approach}


We aim to create a general purpose pre-training architecture that can be trained to produce an effective state-action representation.
Our model is called the Perception-Action Causal Transformer, (\method{}) which works as a causal transformer that ingests perception and action data. Note that, as is typical in robotics, the states that are being learned by our model are not the ground truth states, but rather sensor observations.
Using this data, \method{} is expected to learn an effective joint representation of states and actions, resulting in an implicit understanding of dynamics in an end-to-end manner. 

\subsection{Tokenization} 
\label{sec:tokenization}

In a large variety of applications, raw observations can be of distinct modalities, (e.g. RGB images, LiDAR scans, depth maps). Similarly, robot actions can be of several types as well such as steering angles, motor commands, or a discrete choice from a predefined library of actions. In order to convert such a wide variety of data into a format that is easily accessible by the Transformer, a tokenization procedure is required. We create state and action tokenizers such that our \method{} model can ingest different observation modalities, and can also handle discrete and continuous action spaces. 
We descrive specific architectures next: \\
\textbf{RGB images:} We use a ResNet-18 backbone \cite{he2016deep} trained from scratch to compute features for RGB images, which are then converted into a token of length $128$.\\
\textbf{Raw LiDAR scans:} We use a 2D LiDAR that returns a sequence of range values. We convert these values into XY locations relative to the vehicle, and then use PointNet \cite{qi2017pointnet} to compute a feature vector. We remove PointNet's transform blocks so that the resulting token is not agnostic to the point cloud's orientation relative to the vehicle.\\
\textbf{BEV LiDAR scans:} An alternate approach to tokenize LiDAR scans was to convert the returns into a bird's eye view (BEV) image of size $200 \times 200$ corresponding to $15 \times 15$ meters around the robot and process it with a ResNet-18 backbone to a token of size $128$. While this technique presented equivalent performance to PointNet in simulation, we found that sim-to-real transfer was more robust using BEV projection. \\
\textbf{Discrete actions:} For Habitat we have a 4D discrete action space with `left', `right', `forward', and `stop' actions. We use a simple linear embedding to map the 4D actions to a token.\\
\textbf{Continuous actions:} We use a 2-layer MLP to map continuous actions into tokens.

\begin{figure}
   \centering
    \includegraphics[width=1.0\columnwidth]{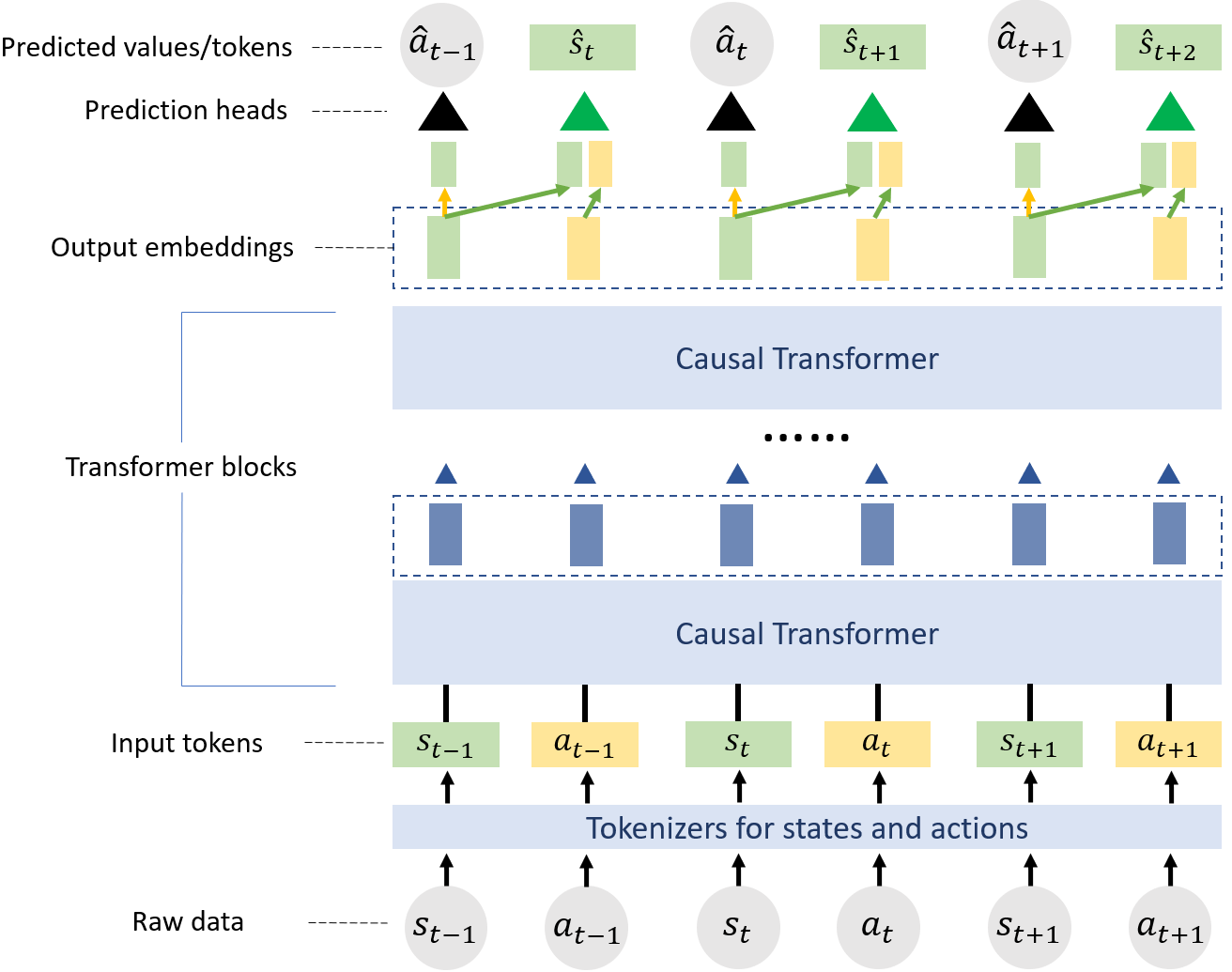}
    \caption{\small{Perception-Action Causal Transformer (PACT) architecture. $\hat{a}$ and $\hat{s}$ are autoregressively predicted actions and states. The tokenizer does not share information across data, and applies operations individually on raw data inputs. The black and green arrows represent predictions heads for actions and future state tokens respectively. }}
    \label{fig:model_arch}
    \vspace{-6mm}
\end{figure}

\subsection{Model design}
\label{sec:model_arch}


The Transformer architecture~\cite{vaswani2017attention} has found significant use in sequence modeling problems. Most Transformer models consist of stacked self-attention layers, and operate on a sequence of embeddings corresponding to input tokens, and transform them into another sequence of embeddings of equal length. 
Let us assume $s_t$ and $a_t$ to represent the state and action at time $t$, and consider a trajectory $\tau$ as a sequence of pairs of the form $\tau = \{(s_0, a_0), (s_1, a_1), ... , (s_T, a_T)\}$.
In our work we attempt to autoregressively model this sequential data of states and actions using the causal transformer architecture, similar to the GPT architecture~\cite{brown2020language}.
The transformer's causal self-attention mask ensures that any particular output token within the sequence is a result of operations over only tokens in past time steps. 

As shown in Fig. \ref{fig:model_arch}, the main blocks in \method{} are the state and action tokenizers, causal transformer blocks and prediction heads.
Our model computes predictions within a limited time horizon of length $K$ which contains a total of $2K$ tokens of alternating states and actions.
Each input is first tokenized into the embedding dimension, and added with a learned positional embedding for each time step. 
Similar to Decision Transformer \cite{chen2021decision}, we use a global time embedding to indicate the token's position within the full sequence ($1 \to 2K$), and a local time embedding which indicates the current time step ($1 \to K$).
This combined sequence of inputs is then fed through multiple layers of causal Transformer blocks to result in a set of output embeddings. 

\subsection{Pre-training Objectives}
Let $q(\cdot)$ represent the tokenizer operation, and $X(\cdot)$ represent the transformer embedding operation for a given token.
To pretrain PACT, we use self-supervised learning with separate action and state prediction heads.
The action prediction head $h_a$ is expected to predict the appropriate next action given the current state embedding $h_a(X(q(s_{t}))) \to a_t$, acting as a policy.
The state prediction head $h_s$ predicts the next state token given the previous state and action embeddings $h_s(\ X(q(s_{t})), X(q(a_{t}))\ ) \to q(s_{t+1})$, acting as a dynamics model.

\subsection{Implementation details}
Model and training parameters are found below:

\begin{table}[h]
\caption{\small{Training parameters for pretraining}}
\footnotesize
\centering
\begin{tabular}{ll}
\toprule
Hyperparameter        & Value \\
\midrule
\# of layers          &   12    \\
\# of attention heads &   8    \\
Embedding length      &   128    \\
Time sequence length       &   16    \\
Batch size            &   32    \\
Pre-training Learning rate         &   6e-4    \\
Finetuning Learning rate         &   6e-5    \\
Weight decay for transformer weights          &   1e-1    \\
Weight decay for all other layers          &   1e-4    \\
LR schedule           &   Ramp-up to 5\%, then decay     \\
Dropout               &   0.1   \\
\bottomrule
\end{tabular}
\end{table}





\section{Downstream Tasks and Experimental Setup}
Our goal is to use the pre-trained PACT representation as a basis for different robotics downstream tasks.
We focus our evaluation on the domain of mobile agents, and show that our robot-specific representations can function as a single starting point towards safe navigation, localization and mapping.

\subsection{Finetuning pipeline}
\label{sec:finetuning}

\begin{figure}
    \centering
    \includegraphics[width=\columnwidth]{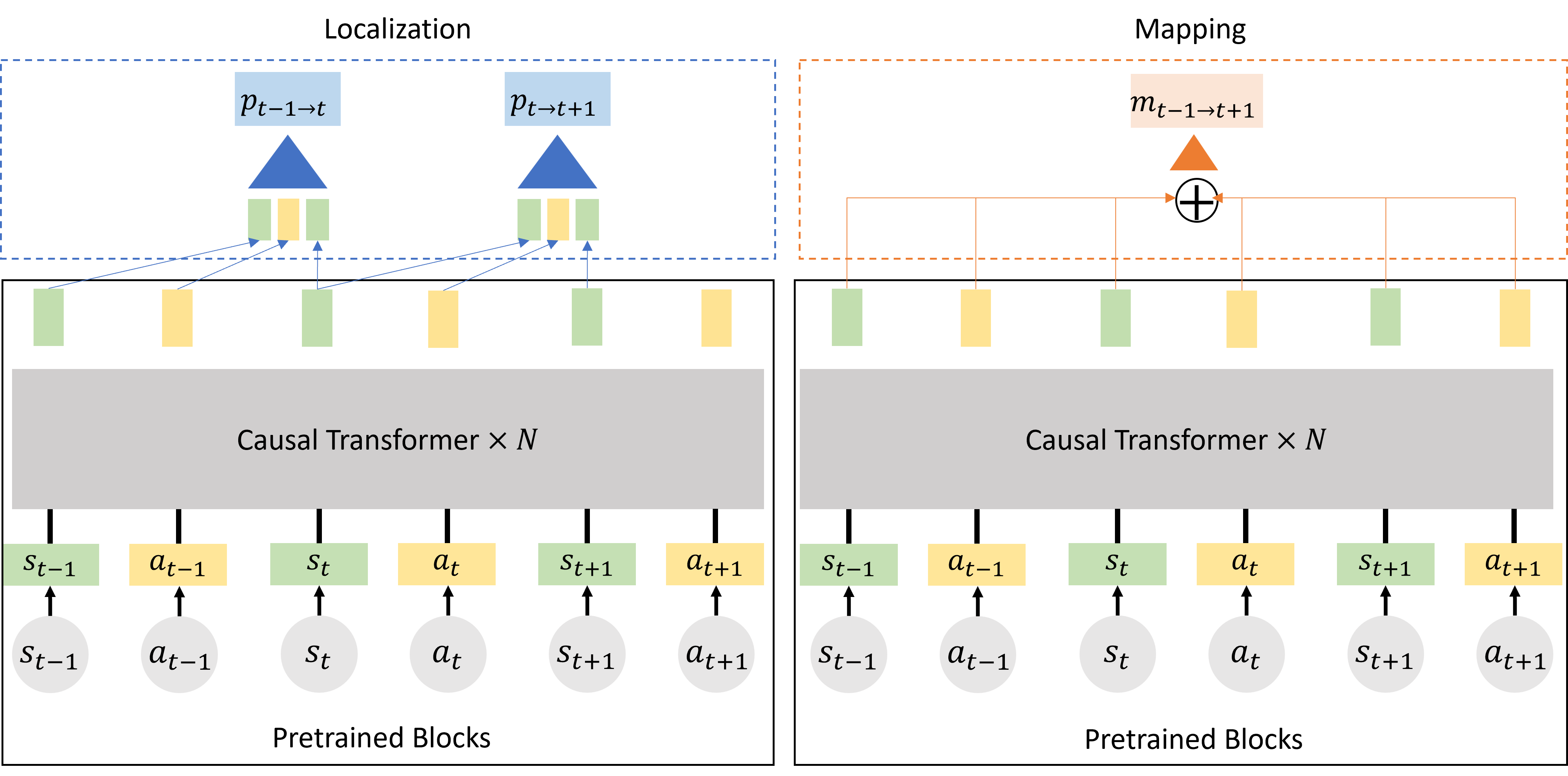}
    \caption{\small{PACT representations can be used for downstream tasks such as localization (left), which needs frame-to-frame computations, and mapping (right) which accumulates data over a window of time.}}
    \label{fig:downstream}
    \vspace{-5mm}
\end{figure}

\textbf{Localization:} Here define robot localization as the task of predicting the robot's pose over time relative to the initial pose of the trajectory. 
Our network runs a deep odometry algorithm, calculating the pose difference between consecutive time steps: $p_{t-1 \to t}$.
We us a lightweight 3-layer MLP $h_{\text{loc}}$ ([$64,32,3$]) as the localization task decoder (Fig. \ref{fig:downstream}, left), whose input is a tuple containing the previous state and action embeddings plus and current state embedding: $p_{t-1 \to t} = h_{\text{loc}}(X(q(s_{t-1})), X(q(a_{t-1})), X(q(s_t)))$.
We train the head using MSE loss with respect to the ground-truth pose difference $\{\Delta X, \Delta Y, \Delta \theta\}$.
The pose differences are integrated over time to estimate the robot's global pose.
Note that we can use the transformer in a rolling-window fashion and accumulate poses over the entire trajectory duration, much longer than the original transformer length $K$.

\textbf{Mapping:} We define mapping as the task of predicting a local 2D occupancy image $I_o$ around the robot's current location (as a top-down view), given the local history (length $K$) of the state and action embeddings.
We use a 2-layer deconvolutional decoder $h_{\text{map}}$ to output an image of size $64\times 64$ that represents the local occupancy of an area of $15\times15$ meters surrounding the robot: $I_o = h_{\text{map}}(X(q(\tau_{j : j+K}))$.
We train the mapping decoder (Fig. \ref{fig:downstream}, right) using an MSE loss over the binary ground-truth pixel values.




\subsection{Experimental setup}


\textbf{MuSHR car:} We apply \method{} to a wheeled robot platform, MuSHR \cite{srinivasa2019mushr}, which is equipped with a 2D LiDAR sensor with angular resolution of $0.5$ degree, and takes a steering angle in the range of $[-21.75, 21.75]$ deg. as control input.
The data for pre-training and downstream tasks is obtained by running a MuSHR simulator on a real-world map collected from an office environment of approximately $75m \times 30m$.
The simulator contains an expert MPC planner which outputs safe trajectories given the occupancy map. 
Using this simulator, we record LiDAR scans and corresponding actions for a total of 5.5K trajectories, or 3M perception-action tokens.

For all simulated experiments the observations, or LiDAR scans, are first converted into a feature vector of size $1024$ using learned PointNet \cite{qi2017pointnet}, and then  reduced into a token of dimension 128. 
For real-world experiments we verified experimentally that processing a top-down LiDAR projection image resulted in better policy transfer, as explained in Section~\ref{sec:tokenization}.
Each action is converted from a continuous scalar into a token in a dimension of 128 using a two-layer MLP.


\begin{figure}
    \centering
    \includegraphics[width=1.0\columnwidth]{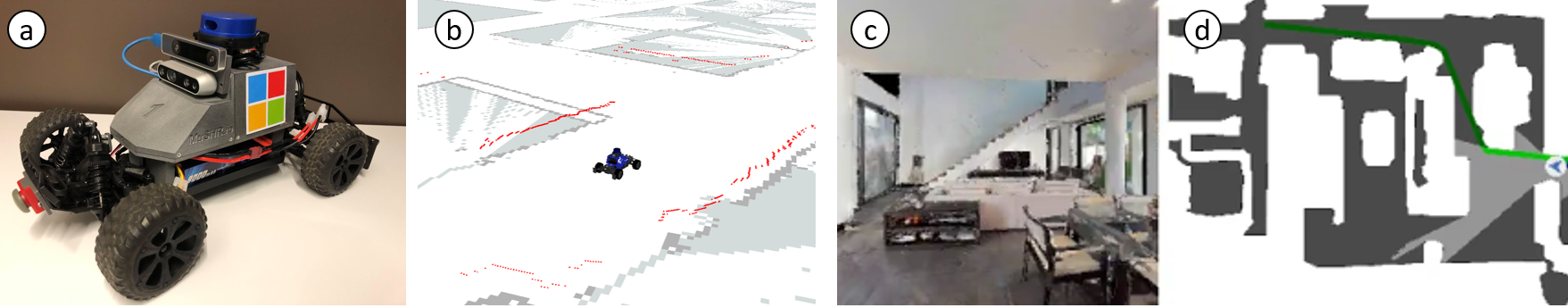}
    \caption{\small{Deployment environments. a) Real-life MuSHR vehicle, b) MuSHR simulator with dynamics model and LiDAR, c) Habitat FPV image, and d) Top-down view of Habitat environment map.}}
    \label{fig:mushr_system}
    \vspace{-5mm}
\end{figure}

\textbf{Habitat:} We also test our method in the Habitat simulator \cite{habitat19iccv}. 
In this case, we train our models on data obtained from expert trajectories on the PointNav task in Habitat. 
We record the first-person camera images and action information across $10$, collecting a total of 10K episodes, or 840K tokens.
Actions belong to a discrete set (`left', `right', `forward', and `stop'). 
We train a ResNet18 \cite{he2016deep} encoder to tokenize the images of size $224\times224$, and use single linear embedding to tokenize the discrete actions from a one-hot vector. 
All embeddings are mapped to a size of $128$. 



\section{Results} 
\label{sec:results}

\subsection{Main Experimental Hypotheses}

We investigate a set of hypotheses regarding pre-training and finetuning representations for navigation scenarios.
The main metrics used here are MSE (mean squared error), MAE (mean absolute error), and ATE (absolute trajectory error).
We detail these hypotheses and results below:

\textbf{H1 - PACT representations can achieve similar or better performance than models trained from scratch:}
As discussed in \cite{he2019rethinking}, there are no theoretical guarantees that finetuning a model starting from a pre-trained representation will necessarily achieve better results than training from scratch, given enough data and time.
In this hypothesis we want to validate that the PACT representation starting point can achieve at least a similar, and ideally better, performance in downstream tasks when compared to training a network from scratch for localization and mapping.

\begin{figure*}
\begin{subfigure}[t]{0.5\textwidth}
         \centering
         \includegraphics[width=\textwidth]{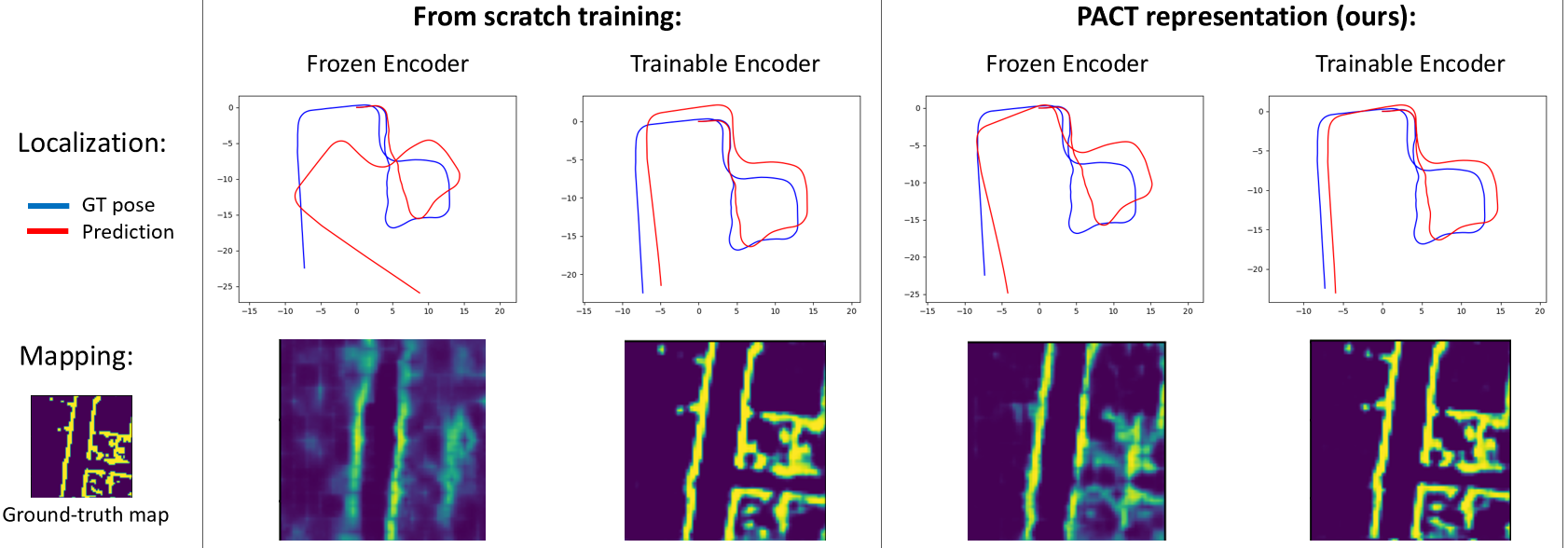}
         \caption{Mushr}
         \label{fig:mushr_downstream}
     \end{subfigure}
     \quad
     \begin{subfigure}[t]{0.5\textwidth}
         \centering
         \includegraphics[width=\textwidth]{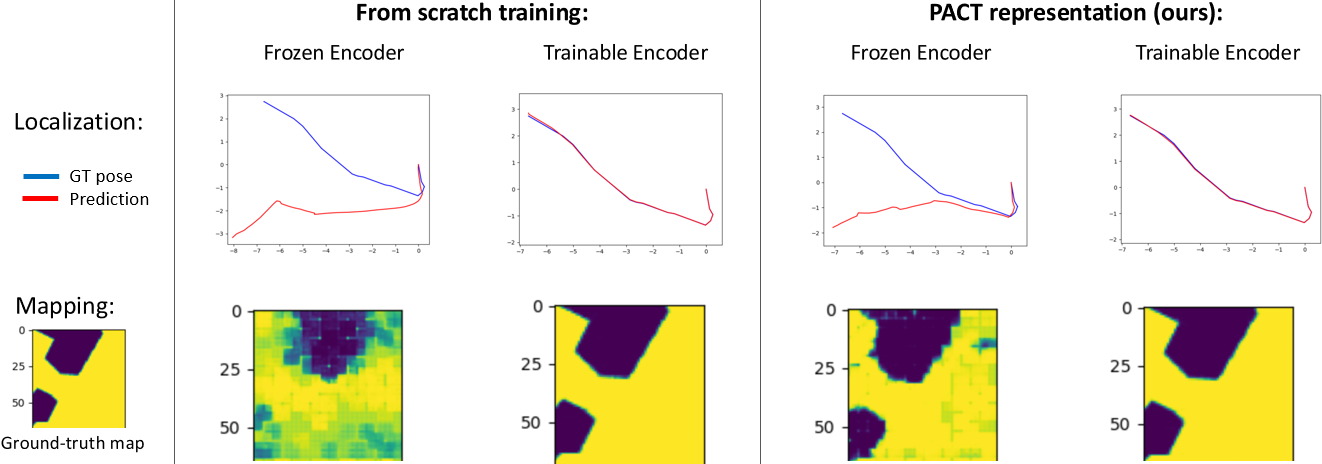}
         \caption{Habitat}
         \label{fig:habitat_downstream}
     \end{subfigure}
     \caption{\small{Visualization of mapping and localization for Mushr (a) and Habitat (b). We compare PACT pre-training against the same model backbone that trained from scratch. Under each setting, we compare frozen and trainable feature encoders.}}
     \label{fig:vis_tasks}
     \vspace{-2mm}
\end{figure*}

Tables \ref{tab:localization_mushr} and \ref{tab:mapping_mushr} show the comparison results. 
For the columns under `PACT' we first pre-train a base model for each environment, and as a second step we finetune this representations for the downstream tasks. 
`F' indicates a frozen representation, where only the small task-specific head is trained, and `T' indicates a fully trainable network.
The `Scratch' column shows networks trained starting from random weights. 
\Cref{fig:vis_tasks} displays visualizations of the output from the four different models for MuSHR and Habitat.


The main conclusions from tables \ref{tab:localization_mushr} and \ref{tab:mapping_mushr} is that in general our hypothesis holds true, and networks derived from PACT pre-training achieve similar and in several cases better results than randomly initialized weights.
We also find, not surprisingly, that trainable representations (T) generally achieve better results than freezing the transformer encoder layers (F). 
The gap in performance between F and T is especially large for the Habitat environment, likely due to the high dimensionality of the image perception modality.

\begin{figure}
    \centering
    \includegraphics[width=1.0\columnwidth]{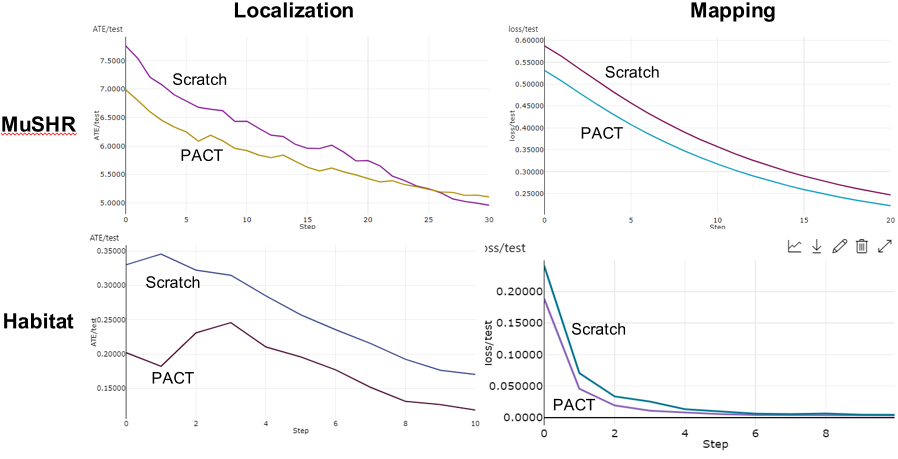}
    \caption{\small{Speed of convergence over training epochs for localization and mapping metrics. We see that pre-training offers a better starting point, even if training from scratch eventually surpasses it. }}
    \label{fig:speed}
    \vspace{-2mm}
\end{figure}

In Figure~\ref{fig:speed} we also investigate the training time required to achieve good performance for pre-trained versus random weights. We can see that PACT offers a better initialization, even if training from scratch eventually surpasses its performance in later epochs. 

\begin{table}[h]
\centering
\caption{\small{Localization ATE averaged over 30 trajectories (meters, lower error is better)}}
\vspace{-4mm}
\begin{subtable}{0.48\columnwidth}
\centering
\caption{MuSHR}
\resizebox{\columnwidth}{!}{%
\begin{tabular}{cc|c|c|c}
    \toprule
    \multirow{2}{*}{\# FT eps.} &
    \multicolumn{2}{c|}{Scratch} & \multicolumn{2}{c}{PACT} \\
    \cline{2-5} \addlinespace
     & F & T & F & T \\ \midrule
100    & 7.13       & 6.30          & 5.51          & \textbf{4.60}             \\
1000   & 5.59       & 4.59          & \textbf{4.21}          & 4.90             \\
5500   & 5.39       & \textbf{3.08}          & 4.09          & 3.62             \\
    \bottomrule
\end{tabular}
}
\end{subtable}
\begin{subtable}{0.48\columnwidth}
\centering
\vspace{2mm}
\caption{Habitat}
\resizebox{\columnwidth}{!}{%
\begin{tabular}{cc|c|c|c}
    \toprule
    \multirow{2}{*}{\# FT eps.} &
    \multicolumn{2}{c|}{Scratch} & \multicolumn{2}{c}{PACT} \\
    \cline{2-5} \addlinespace
     & F & T & F & T \\ \midrule
800   & 1.94   & 0.22     & 2.64     & \textbf{0.16}            \\
2400   & 1.30   & 0.13     & 1.72     & \textbf{0.03}             \\
4000   & 1.27   &  0.08    & 1.47     & \textbf{0.03}              \\
8000   & 1.14   & 0.04     & 1.01     & \textbf{0.02}             \\
    \bottomrule
\end{tabular}
}
\end{subtable}
\label{tab:localization_mushr}
\vspace{-4mm}
\end{table}

\begin{table}[h]
\footnotesize
\centering
\caption{\small{Local map reconstruction performance MSE}}
\begin{subtable}{0.48\columnwidth}
\centering
\caption{MuSHR}
\resizebox{\columnwidth}{!}{%
\begin{tabular}{cc|c|c|c}
    \toprule
    \multirow{2}{*}{\# FT eps.} &
    \multicolumn{2}{c|}{Scratch} & \multicolumn{2}{c}{PACT} \\
    \cline{2-5} \addlinespace
     & F & T & F & T \\ \midrule
100    &  0.566      & 0.425          & \textbf{0.365}          & 0.503             \\
1000   & 0.47       & 0.182          & 0.326          & \textbf{0.167}             \\
5500   & 0.43       & 0.143          & 0.300          & \textbf{0.134}             \\
    \bottomrule
\end{tabular}
}
\end{subtable}
\begin{subtable}{0.48\columnwidth}
\centering
\caption{Habitat}
\resizebox{\columnwidth}{!}{%
\begin{tabular}{cc|c|c|c}
    \toprule
    \multirow{2}{*}{\# FT eps.} &
    \multicolumn{2}{c|}{Scratch} & \multicolumn{2}{c}{PACT} \\
    \cline{2-5} \addlinespace
     & F & T & F & T \\ \midrule
800    & 0.368       & 6.4e-4  & 0.166  & \textbf{5.2e-4}       \\
2400    & 0.326       & \textbf{2.61e-4} & 0.1    & 2.78e-4       \\
4000    & 0.297       & \textbf{1.72e-4} & 0.075  & 1.98e-4   \\
8000    & 0.26        & \textbf{1.56e-4} & 0.053  & 1.78e-4 \\
    \bottomrule
\end{tabular}
}
\end{subtable}
\label{tab:mapping_mushr}
\vspace{-2mm}
\end{table}

\textbf{H2 - The PACT representation is a good starting point for multiple tasks:}
One of our major objectives with \method{} is to show that it can serve as a general common representation, useful for a diverse set of downstream tasks. 
Leveraging a common representation is advantageous for a neural robotics architecture with real-time compute constraints because we can use a single expensive feature extraction module followed by lightweight decoders for each task as opposed to multiple individual large networks.
In addition, the training process for downstream tasks should require less data and compute if initialized with a good representation.

In \Cref{tab:joint_training} we make a direct comparison between multi-head training for localization, mapping and navigation simultaneously from scratch versus initialization with PACT features.
The table shows that PACT features can achieve superior performance for our mobile agent tasks. 
The performance gap holds especially in low-data regimes, showing that our transformer can effectively use the pre-training procedure to serve as a good-quality initialization. 


\begin{table}[h]
\centering
\caption{\small{Joint multi-head training from scratch versus training from frozen PACT representation}}
\begin{subtable}{\columnwidth}
    \scriptsize
\centering
\caption{MuSHR}
\resizebox{\columnwidth}{!}{%
\begin{tabular}{cc|c|c|c}
    \toprule
    \multirow{2}{*}{\# of training episodes} &
    \multicolumn{2}{c|}{Multi-head training from scratch} & \multicolumn{2}{c}{Finetuned \method{} representation} \\
    \cline{2-5} \addlinespace
     & Loc ATE & Map MSE  & Loc ATE & Map MSE  \\ \midrule
100     & 18.5       & 0.59                   & \textbf{10.3}   & \textbf{0.40}                  \\
1000    &  8.12      & 0.31                & \textbf{5.46}   & \textbf{0.25}                     \\
5500    & 4.87       & 0.22                 & \textbf{3.84}   & \textbf{0.21}                  \\
    \bottomrule
\end{tabular}
}
\end{subtable}
\\
\vspace{2mm}
\begin{subtable}{\columnwidth}
    \scriptsize
\centering
\caption{Habitat}
\resizebox{\columnwidth}{!}{%
\begin{tabular}{cc|c|c|c}
    \toprule
    \multirow{2}{*}{\# of training episodes} &
    \multicolumn{2}{c|}{Multi-head training from scratch} & \multicolumn{2}{c}{Finetuned \method{} representation} \\
    \cline{2-5} \addlinespace
     & Loc ATE & Map MSE  & Loc ATE & Map MSE  \\ \midrule
800      & 1.233      & 0.173               & \textbf{0.802}   & \textbf{0.055}                    \\ 
2400     &  0.263      & 0.023                   & \textbf{0.185}   & \textbf{0.010}                   \\
4000    &  0.239      & 0.016                   & \textbf{0.0.126}   & \textbf{0.008}                     \\
8000    &  0.119      & 0.007                  & \textbf{0.105}   & \textbf{0.006}                  \\
    \bottomrule
\end{tabular}
}
\end{subtable}
\label{tab:joint_training}
\end{table}

\textbf{H3 - Both perception and action are important for representation learning:}
As seen in \cref{sec:related_work}, previous methods for robotics representation learning work almost exclusively only with perception features \cite{parisi2022unsurprising,wulfmeier2021representation,wu2022daydreamer}, treating model dynamics as a separate entity.
Our hypothesis is that a feature representation method that combines both perception and action jointly can encode more useful information for downstream tasks.

Similar to some of the analysis in \cite{yang2021representation}, we conduct an ablation study to understand the effect of different auxiliary pre-training losses.
\Cref{tab:pretext_compare} displays metrics for different downstream task decoders trained on different frozen PACT representations: exclusively using state or action prediction losses, or using both simultaneously during pre-training.
We see that representations that include a combination of both state and action prediction losses yields the richest representations for localization and mapping.

\begin{table}[h]
\caption{Pre-training comparison}
\begin{center}
\resizebox{\columnwidth}{!}{%
\begin{tabular}{c|c|c|c|c}
    \toprule
    \# of episodes & Task & State only & Action only & State and Action \\ \midrule
    \multirow{2}{*}{MuSHR} & Localization ATE & 52.0 & 4.34 & \textbf{4.21} \\ 
                            & Mapping MSE & 0.609 & 0.331 & \textbf{0.326} \\     
    \midrule
     \multirow{2}{*}{Habitat} & Localization ATE  &  2.92   & 3.218    & \textbf{1.512}  \\ 
                          & Mapping MSE & 0.596  & 0.110  & \textbf{0.053}   \\                        
    \bottomrule
\end{tabular}
}
\end{center}
\label{tab:pretext_compare}
\end{table}

\subsection{Additional Ablations}

\begin{figure}[h]
    \centering
    \includegraphics[width=1.0\columnwidth]{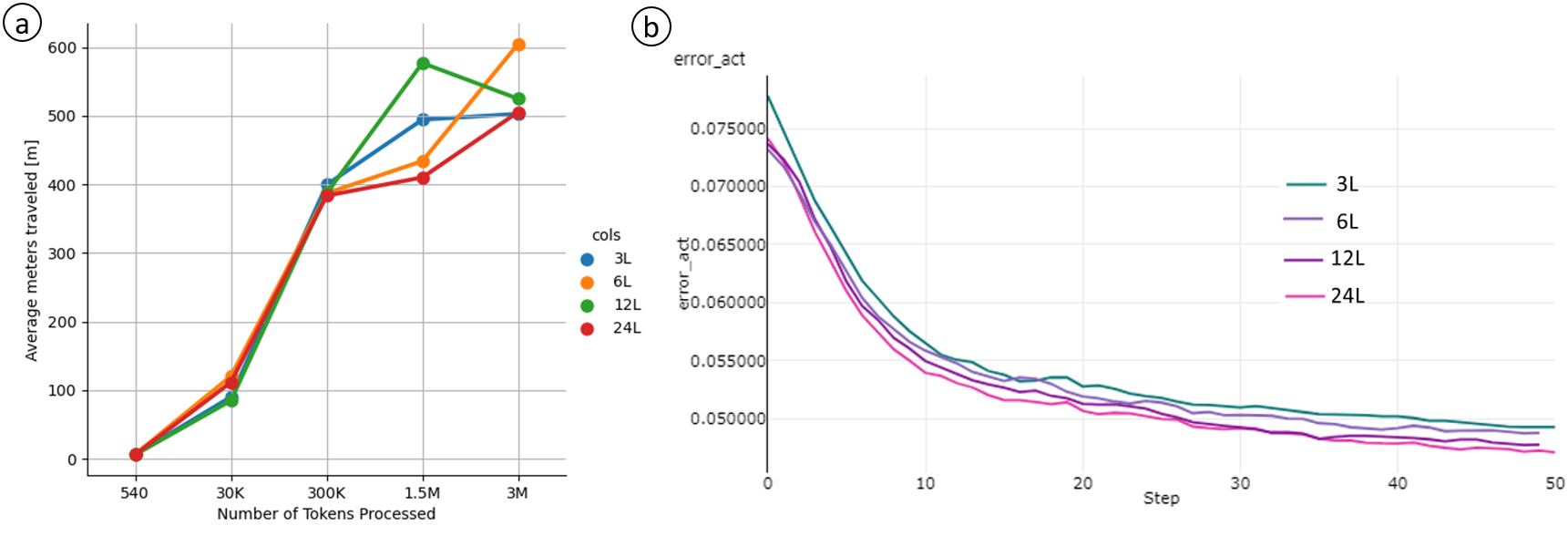}
    \caption{\small a) Effect of model and dataset sizes on pre-training performance measured as the average number of meters traversed until a crash; b) Effect of transformer layer depth on action prediction MAE.}
    \label{fig:model_sizes}
\end{figure}

\textbf{Model size and transformer sequence length:} 
We analyze the pre-training model performance for MuSHR as a function of the number of tokens used for training and as a function of model capacity, expressed as the number of layers of the transformer architecture. We evaluated 4 model sizes (3, 6, 12, 24 layers), as shown in Fig~\ref{fig:model_sizes}a. Performance is measured in terms of average number of meters traversed over 150 model deployments in a realistic floor plan.

In general we see an improvement in model performance as we increase the number of training tokens. Interestingly, larger models did not necessarily result in better performance for robot navigation. Even though larger models consistently presented better loss values for action prediction on a static dataset,
(Fig.~\ref{fig:model_sizes} b), 
when it comes to real-time deployment the larger network capacity introduces inference delays that become a disadvantage and lead to earlier crashes. For example, while LiDAR perception measurements arrive to the vehicle every 0.077s ($13$Hz), the largest model of 24 layers takes on average 0.023s for inference with a RTX3090 GPU, roughly 40\% longer the 3 layer model (0.016s). These time differences can amount to even larger performance gaps in small embedded systems, and further emphasize the importance of multiple downstream task architectures sharing a common representation branch for real-time robotics applications.


\begin{figure}[t]
    \centering
    \includegraphics[width=0.7\columnwidth]{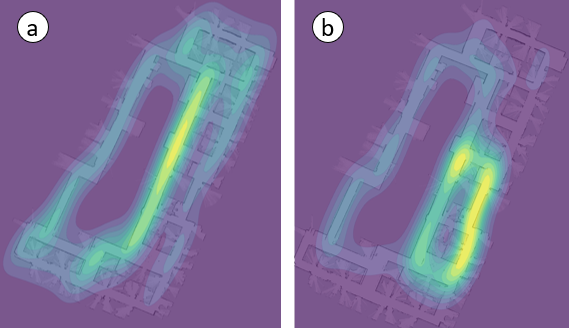}
    \caption{\small{Visualization of state distribution heatmaps for different prompting sequences.}}
    \label{fig:heatmap}
    \vspace{-4mm}
\end{figure}

\textbf{Generative properties of the pre-trained model:}
Analogous to how a model like GPT-3 operates, we can bias the future distribution of states and actions produced by our pre-trained model by prompting the transformer sequence with specific initial values. Fig.~\ref{fig:heatmap} displays heatmaps with state distributions for multiple runs where the car was initialized in the same position and orientation, and the only difference being the prompting of the very first $15$ action tokens. The figure highlights that prompting with straight trajectories results in future actions that tend to keep the vehicle on a straighter course when compared to the actions that are generated from prompts including turns. 
It demonstrates the effectiveness of \method{} to generate sequences of actions that mimic a desired behavioral prompt.

\begin{figure}[h]
    \centering
    \includegraphics[width=1.0\columnwidth]{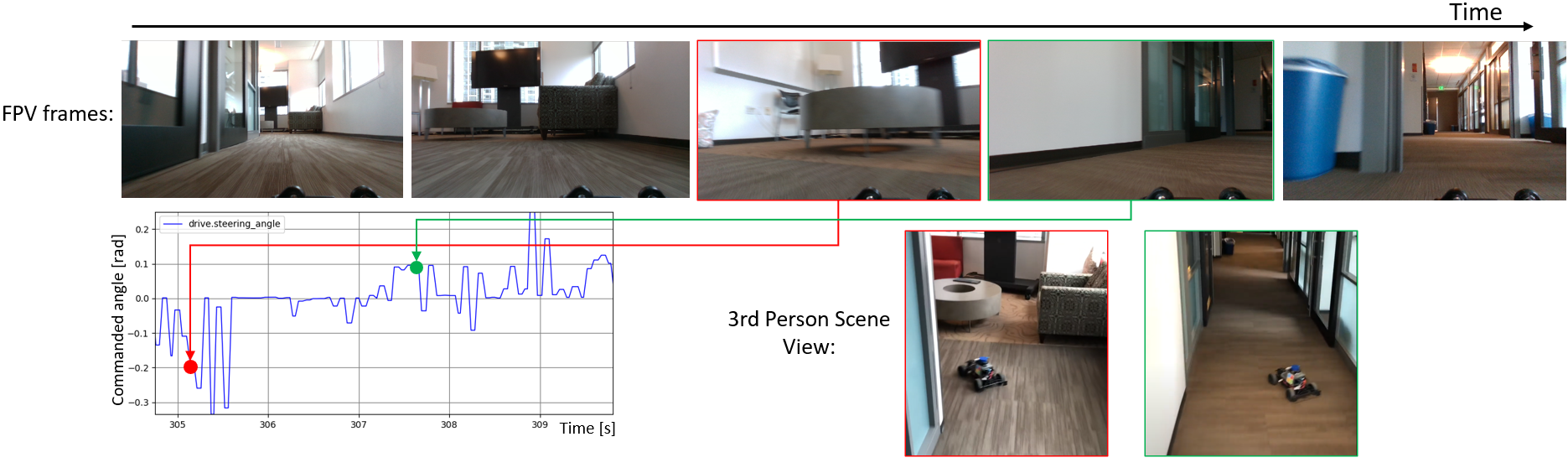}
    \caption{\small{Real-world deployment of pre-trained model}}
    \label{fig:realworld}
\end{figure}

\textbf{Real-world experiments:} Even thought the entirety of the pre-training data was generated in a robotics simulator, we tested how transferable the transformer features and action prediction decoder were when deployed \textit{in the wild}. We deploy the MuSHR car in real life and show results in \Cref{fig:realworld}.
Longest runs we observed were in the order of 80 meters, despite the sim-to-real gap, demonstrating the robustness of our model.

\section{Conclusion and Discussion}
\label{sec:discussion}

We presented PACT, a Perception-Action Causal Transformer architecture aimed at building a general purpose representation from robot data in a self-supervised fashion.
Through autoregressive prediction of states and actions over time, we showed that our model can implicitly encode robot states, dynamics and behaviors. 
Such a representation, built in a robot-specific way, can function as a single starting point to achieve distinct tasks such as safe navigation, localization and mapping. 

We demonstrated our approach in two navigation scenarios using a wheeled robot and a simulated agent, and showed that fine-tuning task-specific networks for localization and mapping on top of the pre-trained models results in better performance compared to models that are trained from scratch.
In addition, sharing a single common representation across tasks is advantageous to
speed up real-time deployment, opening a promising avenue towards foundational models for robotics and control tasks.

When considering future work, we highlight the fact that \method{} does not necessarily require \textit{optimal} demonstrations to learn statistical patterns between state-action tokens, which lowers the burden of developing expert trajectories towards collecting \textit{reasonable} demonstrations. 
We are interested in a more formal analysis of how demonstration quality affects the representation performance.



\footnotesize{
\bibliographystyle{IEEEtran}
\bibliography{IEEEexample.bib}
}


\end{document}


\maketitle

\section{Experimental Details}

\subsection{Dataset Collection}

\textbf{MuSHR:} We adapt the codebase available from the MuSHR open-sourced project (\citet{srinivasa2019mushr}, available at \url{https://mushr.io/tutorials/quickstart/}) towards designing our simulator setup. 
The data collection procedure used a MPC controller to generate a trajectory library with $27$ candidate trajectories at each time step, and the lowest-cost trajectory (considering obstacle avoidance and control effort minimization) was selected at a rate of $50$Hz during re-planning.
The simulation environment consisted of a real office floor plan of approximately $74m \times 30m$ which was mapped using the \textit{gmapping} library \citep{grisetti2007improved}, and for each episode we sampled a random valid goal location.
In total we collected approximately $1.5$ million perception-action pairs of the vehicle in action. The data consisted of 2D LiDAR measurements with angular resolution of $0.5$ degrees ($720$ returns per scan), and vehicle wheel angle (limited to a motion amplitude of of 43.5 degrees).

\textbf{Habitat:} We use the Habitat simulator \citep{habitat19iccv} and sample random valid goal locations for the agent accross $10$ environments. Use use Habitat's built-in shortest path function to generate the agent's actions, and record a total of $800$K perception-action pairs consisting of RGB images of size $224 \times 224$ with their respective discrete actions (left turn, right turn, go forward, stop).


\subsection{Tokenizer Network Architectures}


\textbf{RGB images.} We use a ResNet-18 backbone \cite{he2016deep} to compute features for RGB images, which are then converted into a token of length 128.

\textbf{PointNet LiDAR scans.} We use a 2D LiDAR that returns a sequence of range values. We convert these values into XY locations in a vehicle-oriented bird's eye view, and then use the PointNet~\citep{qi2017pointnet} to compute a feature for each scan, which is then converted into a token. We remove PointNet's transform blocks so that the resulting token is not agnostic to the point cloud's orientation relative to the vehicle. 

\textbf{BEV LiDAR scans.} For real-world experiments only we found that using a ResNet-18 backbone was more robust as a tokenizer for LiDAR data. There, we converted LiDAR return values into a bird's eye view image of size $200 \times 200$, which is processed through a ResNet-18 backbone, results and finally converted into a token of length of size $128$.

\textbf{Discrete actions.} In our experiment on Habitat, we have a 4-D discrete action space, i.e. `left', `right', `forward', and `stop'. To tokenize such a discrete action space, we use a simple linear embedding to map the 4-D actions to a token.

\textbf{Continuous actions.} In our experiments with a continuous action space we are use a 2-layer MLP to map actions into a 128-D token embedding. 


\subsection{Training Parameters}



The training and network parameters used for the main paper experiments are described in the table below, unless where noted differently.

\begin{table}[h]
\footnotesize
\centering
\begin{tabular}{ll}
\toprule
Hyperparameter        & Value \\
\midrule
\# of layers          &   12    \\
\# of attention heads &   8    \\
Embedding length      &   128    \\
Sequence length       &   16    \\
Batch size            &   32    \\
Pre-training Learning rate         &   6e-4    \\
Finetuning Learning rate         &   6e-5    \\
Weight decay for transformer weights          &   1e-1    \\
Weight decay for all other layers          &   1e-4    \\
LR schedule           &   Ramp-up (5\% of total tokens) followed by decay     \\
Dropout               &   0.1   \\
\bottomrule
\end{tabular}
\end{table}



\section{Additional Training Results}

\subsection{Model size and dataset size for MuSHR}

\begin{figure}[h]
    \centering
    \includegraphics[width=0.7\textwidth]{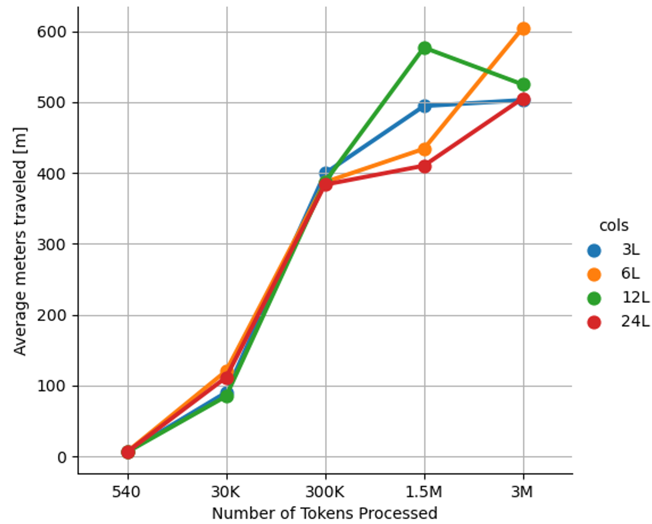}
    \caption{\small Effect of model and dataset sizes on pre-training performance. Performance is measured as the average number of meters traversed until a crash for each model during deployments.}
    \label{fig:model_sizes}
\end{figure}

We analyze the pre-training model performance for MuSHR as a function of the number of tokens used for training and as a function of model capacity, expressed as the number of layers of the transformer architecture. We evaluated 4 model sizes (3, 6, 12, 24 layers), as shown in Fig~\ref{fig:model_sizes}. Performance is measured in terms of average number of meters traversed over 150 model deployments in a realistic floor plan.

\begin{figure}[h]
    \centering
    \includegraphics[width=0.7\textwidth]{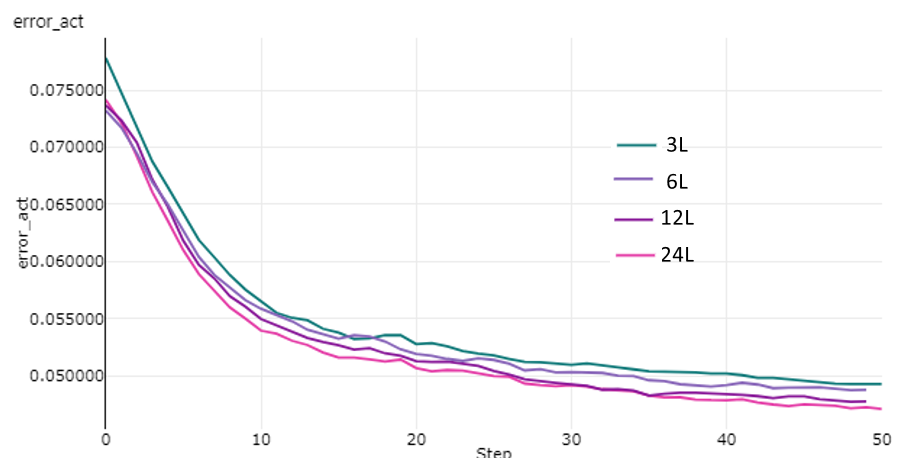}
    \caption{\small Effect of model sizes on pre-training action prediction mean absolute error for each training epoch. All models trained on 1.5M tokens.}
    \label{fig:model_sizes_act}
\end{figure}

In general see an improvement in model performance as we increase the number of training tokens. Interestingly, larger models did not necessarily result in better performance for robot navigation. Even though larger models consistently presented better loss values for action prediction on a static dataset (Fig.~\ref{fig:model_sizes_act}), when it comes to real-time deployment the larger network capacity introduces inference delays that become a disadvantage and lead to earlier crashes. For example, while LiDAR perception measurements arrive to the vehicle every 0.077s ($13$Hz), the largest model of 24 layers takes on average 0.023s for inference with a RTX3090 GPU, roughly 40\% longer the 3 layer model (0.016s). These time differences can amount to even larger performance gaps in small embedded systems, and further emphasize the importance of multiple downstream task architectures sharing a common representation branch for real-time robotics applications.

\subsection{Attention Maps}

\begin{figure}[h]
    \centering
    \includegraphics[width=1.0\textwidth]{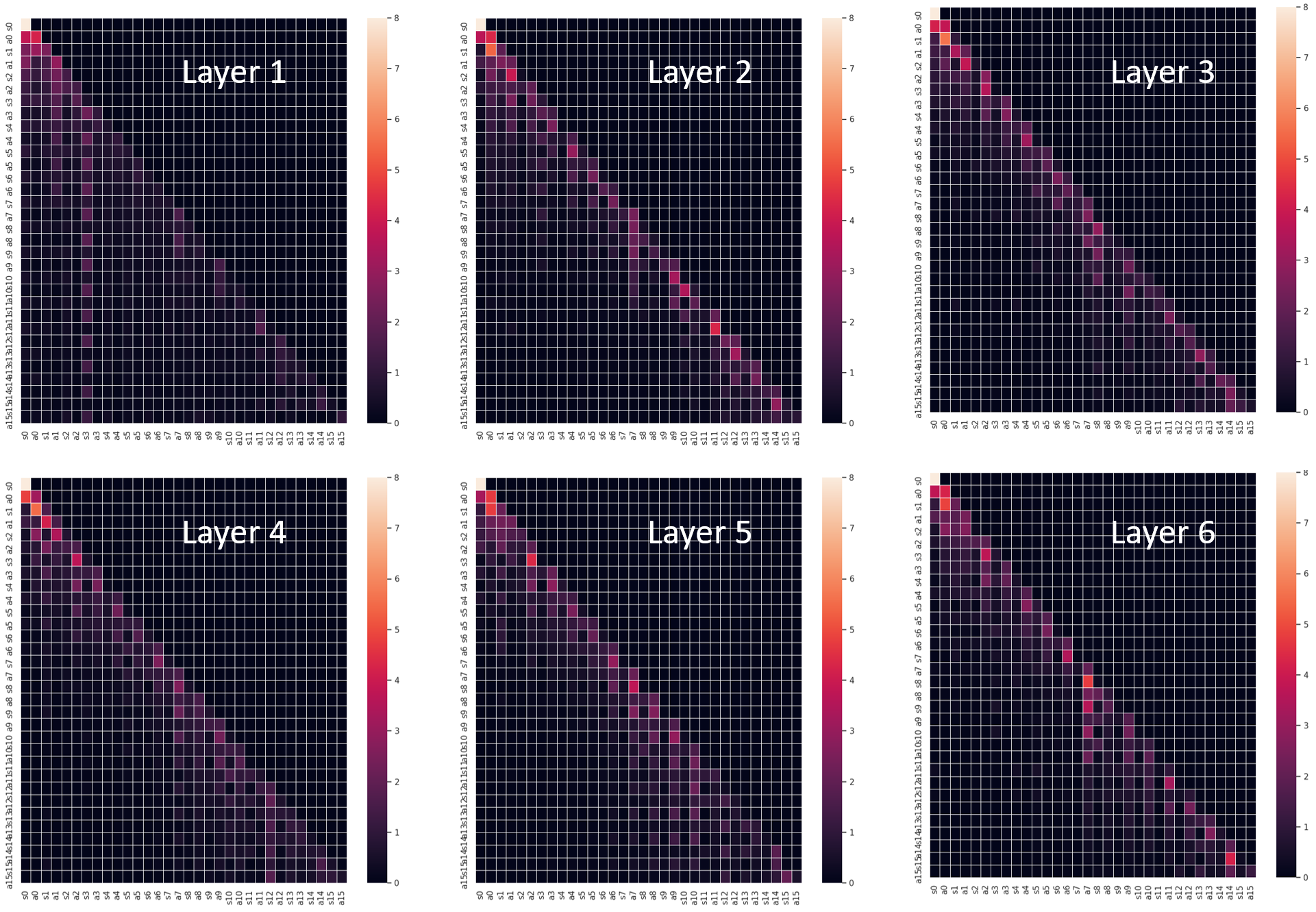}
    \caption{Visualization of attention map for the first $6$ layers (out of 12) of the transformer for MuSHR, summed over 8 heads. Different layers might learn different concepts and be more or less focused on particular significant time steps in the past.Notice that for this particular example all actions after $s_3$ have high attention values towards $s3$ for the first layer, but attention gets more distributed in upper layers. }
    \label{fig:vis_att}
\end{figure}

\begin{figure}[h]
    \centering
    \includegraphics[width=1.0\textwidth]{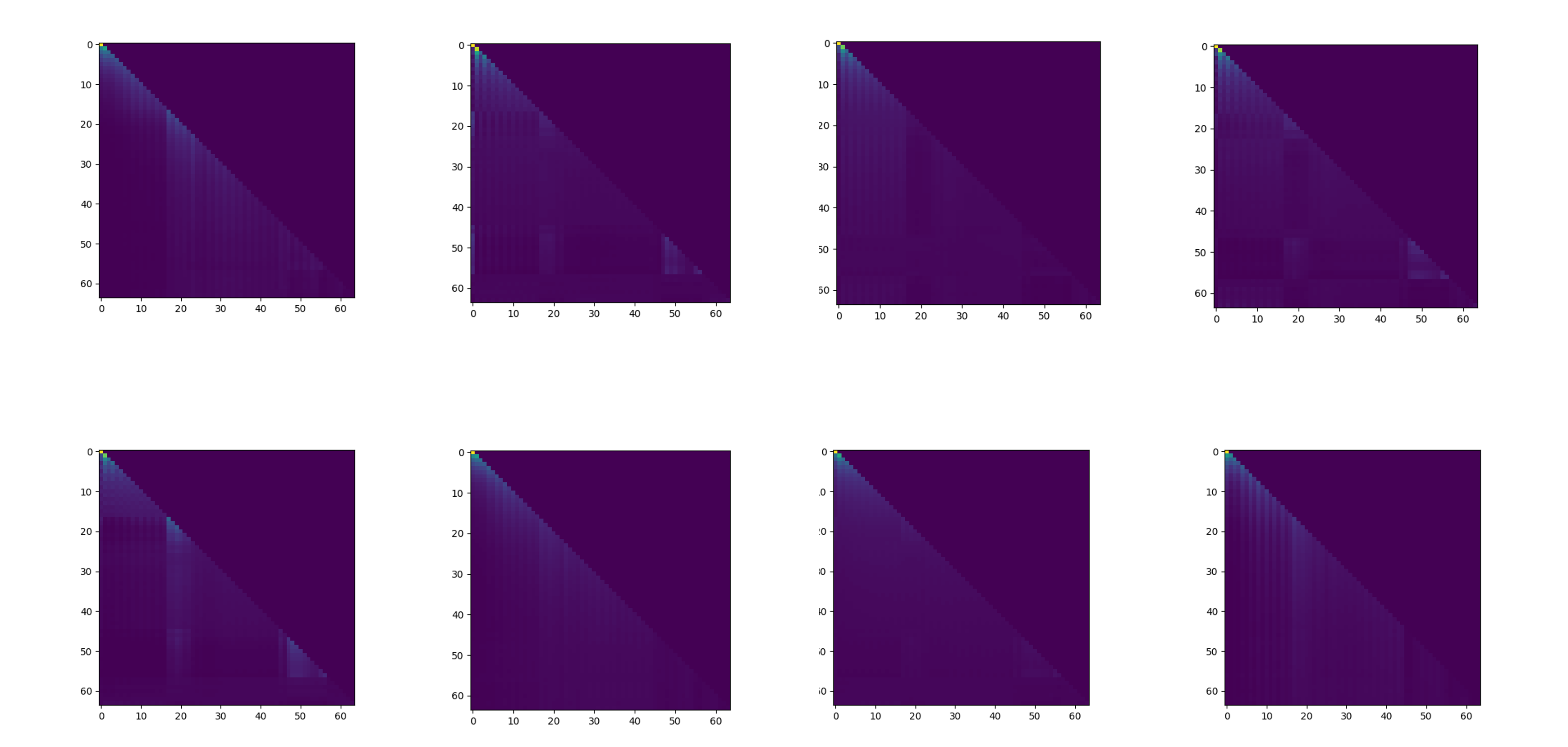}
    \caption{Visualization of the learned attention maps for different heads in the last layer of pretrained PACT on Habitat. As we can see, different attention heads learn to capture different dynamic patterns from the query. For example, some heads learned to attend more on the starting point of an episode, while some others attend more the state change points.}
    \label{fig:my_label}
\end{figure}

We visualize the attention maps for MuSHR (Fig.~\ref{fig:vis_att}) and Habitat (Fig.~\ref{fig:my_label}).
For both maps we have states and actions intercalated in time order ($s_0, a_0, s_1, a_1, ...$). 
To interpret the attention maps one can consider that the embedding in row $i$ \textit{pays attention to} the embedding in column $j$. Matrices are lower-diagonal because of the causal transformer architecture, where tokens at time $t$ can only attend to tokens from the beginning of the sequence up until that step.

\subsection{Sequence length and accuracy}

\begin{figure}[h]
    \centering
    \includegraphics[width=0.6\textwidth]{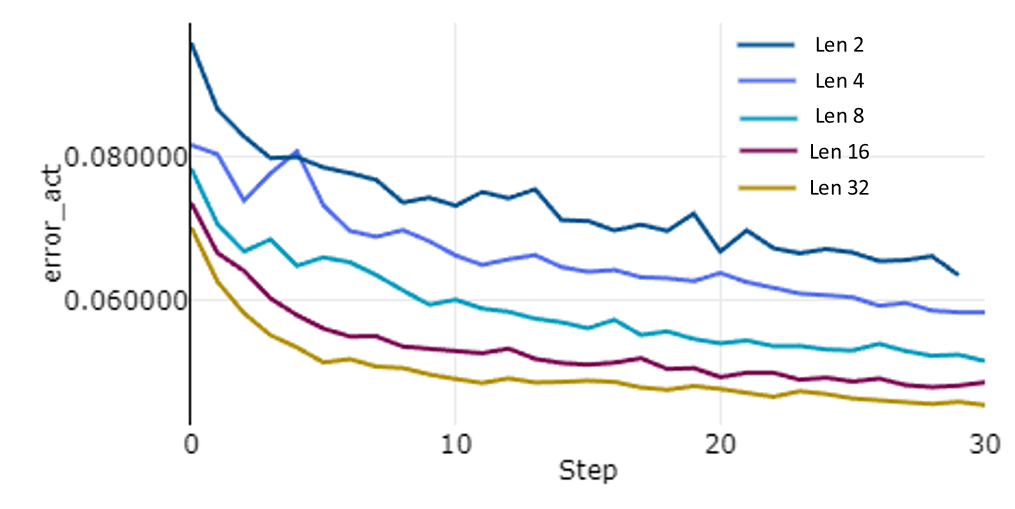}
    \caption{Visualization of how the transformer sequence length affects action prediction mean absolute error (MAE). X axis represents the training epoch number, and Y axis shows the action prediction MAE. We can see that longer sequences translate to better predictions.}
    \label{fig:seq_len}
\end{figure}

We evaluate the impact of longer transformer sequence lengths on the accuracy of action prediction in the pre-trained PACT model.
As seen in Figure~\ref{fig:seq_len}, longer sequences lead to lower mean absolute errors of action prediction.
In practice one must find a good trade-off point because longer sequences lead to longer model training time, and signific longer delays in real-time deployments.
For our main paper experiments we used a sequence length of size $16$, which presented a good trade-off between accuracy and real-time performance.

\subsection{Habitat downstream tasks}
\begin{figure}
    \centering
    \includegraphics[width=1.0\textwidth]{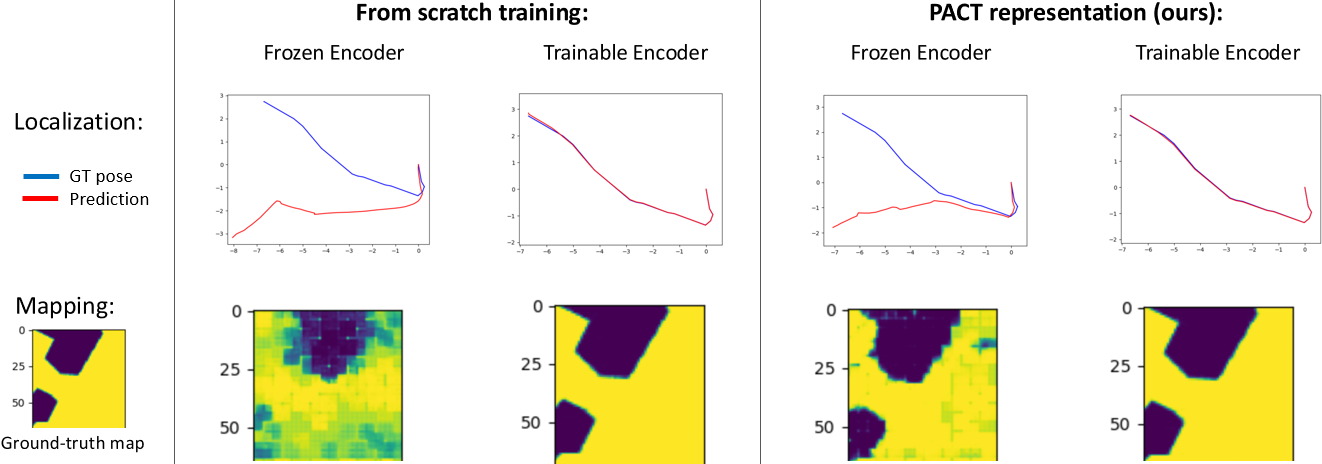}
    \caption{Visualization of Habitat downstream tasks, comparing results obtained from frozen and trainable representations trained from scratch and from PACT.}
    \label{fig:habitat_down}
\end{figure}

We present additional visualizations of Habitat's downstream tasks of mapping and localization in Figure~\ref{fig:habitat_down}.

\newpage
\bibliography{example}  